\documentclass[11pt,letterpaper]{article}
\usepackage{emnlp2016}
\usepackage{times}
\usepackage{latexsym}
\usepackage{amsmath}
\DeclareMathOperator*{\argmax}{arg\,max}
\usepackage{graphicx}
\usepackage{url}
\usepackage{multirow}
\usepackage{enumitem}
\setlist{nolistsep,leftmargin=*}
\emnlpfinalcopy

\def\emnlppaperid{943}

\title{Characterizing the Language of Online Communities and \\
its Relation to Community Reception}

\author{Trang Tran \and Mari Ostendorf \\
Electrical Engineering, University of Washington, Seattle, WA \\
{\tt \{ttmt001,ostendor\}@uw.edu}}

\date{}

\begin{document}
\maketitle

\begin{abstract}
This work investigates style and topic aspects of language in online communities: looking at both utility as an identifier of the community and correlation with community reception of content. Style is characterized using a hybrid word and part-of-speech tag n-gram language model, while topic is represented using Latent Dirichlet Allocation. Experiments with several Reddit forums show that style is a better indicator of community identity than topic, even for communities organized around specific topics. Further, there is a positive correlation between the community reception to a contribution and the style similarity to that community, but not so for topic similarity. 
\end{abstract}

\section{Introduction}
\label{sec:intro}
Online discussion forums provide a rich source of data for studying people's language usage patterns. Discussion platforms take on various forms: articles on many news sites have a comment section, many websites are dedicated to question answering (\url{www.quora.com}), and other platforms let users share personal stories, news, and random discoveries (\url{www.reddit.com}).  Like their offline counterparts, online communities are often comprised of people with similar interests and opinions. Online communication, however, differs from in-person communication in an interesting aspect: explicit and quantifiable feedback. Many discussion forums give their users the ability to upvote and/or downvote content posted by another user. These explicit reward/penalty labels provide valuable information on the reaction of users in a community. In this work, we take advantage of the available user response to explore the relationship between community reception and the degree of stylistic/topical coherence to such communities. Using hybrid n-grams and Latent Dirichlet Allocation (LDA) topic models to represent style and topic for a series of classification tasks, we confirm that there exists a \emph{community language style}, which is not simply characterized by the \emph{topics} that online communities are organized around. Moreover, we show that language style is better at discriminating communities, especially between different communities that happen to discuss similar issues. In addition, we found a positive, statistically significant, correlation between the community feedback to comments and their style, but  interestingly not with their topic. Finally, we analyze community language on the user level and show that more successful users (in terms of positive community reception) tend to be more \emph{specialized}; in other words, analogous to offline communities, it is rare for a person to be an expert in multiple areas. 
%

\section{Related Work}
\label{sec:relatedwork}
It is well known that conversation partners become more linguistically similar to each other as their dialogue evolves, via many aspects such as lexical, syntactic, as well as acoustic characteristics \cite{Niederhoffer2002,Levitan2011}. This pattern is observed even when the conversation is fictional \cite{Danescu2011}, or happening on social media \cite{Danescu2011b}. Regarding the language of online discussions in particular, it has been shown that individual users' linguistic patterns evolve to match those of the community they participate in, reaching ``linguistic maturity'' over time  \cite{Nguyen2011,Danescu2013}. In a multi-community setting, Tan and Lee \shortcite{Tan2015} found that users tend to explore more in different communities as they mature, adopting the language of these new communities. These works have mainly focused on the temporal evolution of users' language. Our work differs in that we use different language models to explore the role of topic and style, while also considering users in multiple communities. In addition, we look at community language in terms of its correlation with reception of posted content.

Other researchers have looked at the role of language in combination with other factors in Reddit community reception. Lakkaraju et al. \shortcite{Lakkaraju2013} proposed a community model to predict the popularity of a resubmitted content, revealing that its title plays a substantial role.
Jaech et al. \shortcite{Jaech2015} considered timing and a variety of language features in ranking comments for popularity, finding significant differences across different communities. In our work, we focus on community language, but explore different models to account for it. 
\section{Data}
\label{sec:data}
Reddit is a popular forum with thousands of sub-communities known as \emph{subreddits}, each of which has a specific theme. We will refer to \emph{subreddits} and \emph{communities} interchangeably. 
Redditors can submit content to initiate a discussion thread whose root text we will refer to as a \emph{post}. Under each \emph{post}, users can discuss the post by contributing a \emph{comment}. Both posts and comments can be upvoted and downvoted, and the net feedback is referred to as \emph{karma} points.

We use eight subreddits that reflect Reddit's diverse topics, while limiting the amount of data to a reasonable size.  In addition, we create an artificial distractor {\small\sf merged\_others} that serves as an open class in our classification tasks and for normalizing scores in correlation analysis. Statistics are listed in Table \ref{tab:data-stats-all}.
The 
{\small\sf merged\_others} set includes 9 other subreddits that are similar in size and content diversity to the previous ones:  
{\small\sf books, chicago, nyc, seattle, explainlikeimfive, science, running, nfl,} and {\small\sf todayilearned.}
Among these extra subreddits, the smallest in size is 
{\small\sf nyc} (1.5M tokens, 76K comments), and the largest is 
{\small\sf todayilearned} (88M tokens, 5M comments).  All data is from the period between January 1, 2014 and January 31, 2015. In each subreddit, 20\% of the threads are held out for testing.

We use discussion threads with at least 100 comments, hypothesizing that smaller threads will not elicit enough community personality for our study. (Virtually all threads kept had only upvotes.) 
For training our models, we also exclude individual comments with non-positive karma ($k\le 0$) in order to learn only from content that is less likely to be downvoted by the Reddit communities; percentages are noted in Table~\ref{tab:data-stats-all}.

\begin{table}[]
\begin{tabular}{l|r|r|r}
subreddit & \multicolumn{1}{c|}{\# posts} & \multicolumn{1}{c}{\# comments} & \multicolumn{1}{|c}{\% $k\le 0$} \\ 
\hline \hline
 askmen & 4.5K & 1.1M & 10.6\\ 
 askscience & 0.9K & 0.3M & 9.1\\
 askwomen & 3.6K & 0.8M & 7.5\\
 atheism & 3.1K & 1.0M & 15.2\\
 changemyview & 2.3K & 0.5M & 16.7\\
 fitness & 2.4K & 0.9M & 8.6\\
 politics & 4.9K & 2.2M & 20.8 \\
 worldnews & 9.9K & 6.0M & 23.6\\ 
 merged\_others & 28.0K & 14.2M & 13.2
\end{tabular}
\centering
\caption{Reddit dataset statistics}
\label{tab:data-stats-all}
\end{table}
%
%
%
%
\section{Models}
\label{sec:models}
We wish to characterize community language via \emph{style} and \emph{topic}. For modeling style, a popular approach has been combining the selected words with part-of-speech (POS) tags to construct models for genre detection \cite{Stamatatos2000,Feldman+09,Bergsma2012} and data selection \cite{Iyer1999,Axelrod2014}. For topic, a common approach is Latent Dirichlet Allocation (LDA) \cite{Blei2003}. We follow such approaches in our work, acknowledging the challenge of completely separating style/genre and topic factors raised previously
\cite{Iyer1999,Sarawgi2011,PetrenzWeb11,Axelrod2014}, which also comes out in our analysis. 
Generative language models are used for characterizing both style and topic, since they are well suited to handling texts of widely varying lengths.

\subsection{Representing Style}
\label{ssec:style}
Replacing words with POS tags reduces the possibility that the style model is learning topic,
but replacing too many words loses useful community jargon. To explore this tradeoff, 
we compared four trigram language models representing different uses of words vs.\ POS tags in the vocabulary:
\begin{itemize}
\item \verb+word_only+: a regular token-based language model (vocabulary: 156K words) 
\item \verb+hyb-15k+: a hybrid word-POS language model over a vocabulary of 15K most frequent words across all communities in our data; all other words are converted to POS tags (vocabulary: 15K words + 38 tags)
\item \verb+hyb-500.30+:  a hybrid word-POS language model over a vocabulary of 500 most frequent words in a subset of data balanced across communities, combined with the union of the 30 next most common words from each of the 17 subreddits; all other words are converted to POS tags (vocabulary: 854 words + 38 tags)
\item \verb+tag_only+: a language model using only POS tags as its vocabulary (vocabulary: 38 tags) 
\end{itemize}
The hybrid models represent two intermediate sample points between the extremes of word-only and tag-only n-grams. For the \verb+hyb-500.30+ model, the mix of general and community-specific words was designed to capture distinctive community jargon. The general words include punctuation, function words, and words that are common in many subreddits (e.g., {\sl sex, culture, see, dumb, simply}). The subreddit-specific words seem to reflect both broad topical themes and jargon or style words, as in (themes vs.\ style/jargon):\\
{\small\sf askmen:} {\sl wife, single} \ vs.\ {\sl whatever, interested}\\
{\small\sf askwomen:} {\sl mom, husband}\ vs.\ {\sl especially, totally}\\
{\small\sf askscience:} {\sl particle, planet} \ vs.\ {\sl basically, x}\\
{\small\sf fitness:} {\sl exercises, muscles} \ vs.\ {\sl cardio, reps, rack}

Tokenization and tagging are done using Stanford coreNLP \cite{corenlp}. 
Punctuation is separated from the words and treated as a word.
All language models are trigrams trained using the SRILM toolkit \cite{Stolcke2002}; modified Kneser-Ney smoothing is applied to the \verb+word_only+ language model, while Witten-Bell smoothing is applied to the \verb+tag_only+ and both hybrid models. 

\subsection{Representing Topic}
\label{ssec:topic}
We train 100- and 200-dimensional LDA topic models \cite{Blei2003} using \verb+gensim+ \cite{gensim}.  We remove all stopwords (250 words) and use tf-idf normalized word counts in each comment (as documents). The vocabulary consists of 156K words, similar to the vocabulary of the \verb+word_only+ language model.
The topic models were trained on a subset of the training data, using all collected subreddits but randomly excluding roughly 15\% of the training data of larger subreddits 
{\small\sf worldnews}, {\small\sf todayilearned}, and {\small\sf nfl}.

The topics learned exhibit a combination of ones that reflect general characteristics of online discussions or topics that arise in many forums, some that have more specific topics, and others that do not seem particularly coherent.
Topics (from LDA-100) that consistently have high probability in all subreddits are shown in Table~\ref{tab:topic-words} with their top 10 words by frequency (normalized by the topic average). 
Topic 19 is likely capturing Reddit's writing conventions and formatting rules.
Broadly used topics reflect women's issues (29) and news events (32, 34). 

\begin{table}[]
\centering
\begin{tabular}{l|l}
ID & \multicolumn{1}{c}{Frequent words} \\ \hline \hline
19 & \begin{tabular}[c]{@{}l@{}}-lsb-, -rsb-, -rrb-, -lrb-, **, reddit, \\ comment, confirmed, spanish, fair\end{tabular} \\ \hline
29 & \begin{tabular}[c]{@{}l@{}}sex, pilots, child, women, abortion, \\ mail, birth, want, episodes, children\end{tabular}\\ \hline
32 & \begin{tabular}[c]{@{}l@{}}tax, government, taxes, iraq, pay, cia, \\ land, money, income, people\end{tabular} \\ \hline
34 & \begin{tabular}[c]{@{}l@{}}africa, war, nation, global, germans, \\ rebels, corruption, nations, fuel, world\end{tabular} 
\end{tabular}
\caption{Examples of broadly used topics.}
\label{tab:topic-words}
\end{table}

Online communities are typically organized around a common theme, but multiple topics might fall under that theme, particularly since some of the ``topics'' actually reflect style. A subreddit as a whole is characterized by a distribution of topics as learned via LDA, but any particular discussion thread would not necessarily reflect the full distribution. Therefore, we characterize each subreddit with multiple topic vectors. Specifically, we compute LDA topic vectors for each discussion thread in a subreddit, and learn 50 representative topic vectors for each subreddit via k-means clustering.

\section{Community Classification}
\label{sec:community-classification}
One method for exploring the relative importance of topic vs.\ style in online communication is through community classification experiments: given a discussion thread (or a user's comments), can we identify the community that it comes from more easily using style characteristics or topic characteristics? 
We formulate this task as a multi-class classification problem (8 communities and ``other''), where samples 
are either at the discussion thread level or the user level. At the thread level, all comments (from multiple people) and the post in a discussion thread are aggregated and treated as a document to be classified.  
At the user level, we aggregate all comments made by a user in a certain subreddit 
and treat the collection (which may reflect multiple topics) as a document to be classified. 

We classify document $d_i$ to a subreddit according to $\hat{j} = \argmax_{j} s_{i,j}$, where $s_{i,j}$ is a score of the similarity of $d_i$ to community $j$. For the style models, $s_{i,j}$ is the log-probability under the respective trigram language model of community $j$. For the topic model, $s_{i,j}$ is computed using $d_i$'s topic vector $v_i$ as follows. For a subreddit $j$, we compute the cosine similarities $sim_{j,k}$ between $v_i$ and the subreddit's topic vectors $w_{j,k}$ for $k=1,\ldots , 50$. The final topic similarity score $s_{i,j}$ is the mean of the top 3 highest similarities: $s_{i,j} = (sim_{j,[1]} + sim_{j,[2]} + sim_{j,[3]})/3$, where $[\cdot]$ denotes the sorted cosine similarities' indices. The top-3 average captures the most prominent subreddit topics (as in a nearest-neighbor classifier).
Averaging over all 50 $sim_{j,k}$ is ill suited to subreddits with broad topic coverage, and leads to poor classification results.

Table \ref{tab:classification-summary} summarizes the community classification results (as average accuracy across all subreddits) for each model described in Section \ref{sec:models}. While all models beat the random baseline of 11\%, the poor performance of the \verb+tag_only+ model confirms that POS tags alone are insufficient to characterize the community. Both for classifying threads and authors, 
\verb+hyb-500.30+ yields the best average classification accuracy, due to its ability to generalize POS structure while covering sufficient lexical content
to capture the community's jargon and key topical themes. 
Neither topic model beats \verb+hyb-500.30+, indicating that topic alone is not discriminative enough for community identification, even though specific communities coalesce around certain common topics.  
The \verb+word_only+ and \verb+hyb-15k+ models have performance on the threads that is similar to the topic models, since word features are sensitive to topic, as shown in \cite{PetrenzWeb11}. 

Classifying authors is harder than classifying threads. Two factors are likely to contribute. First, treating a whole discussion thread as a document yields more data to base the decision on than a collection of author comments, since there are many authors who only post a few comments. Second, authors that have multi-community involvement may be less adapted to a specific community. 
The fact that word-based style models outperform topic models may be because the comments are from different threads so not matching typical topic distributions. 

\begin{table}[]
\begin{tabular}{l|c|c}
Model & by thread & by author \\ \hline \hline 
random & 11.1\% & 11.1\% \\ \hline
word\_only & 68.9\% & 46.8\% \\
tags\_only & 27.6\% & 18.8\% \\
hyb-15k & 69.4\% & 46.6\% \\
hyb-500.30 & \textbf{86.5\%} & \textbf{51.0\%} \\ \hline
topic-100 & \textbf{71.1\%} & 27.5\% \\
topic-200 & 69.6\% & \textbf{27.7\%} \\
\end{tabular}
\centering
\caption{Average accuracy for classifying by posts and authors}
\label{tab:classification-summary}
\end{table}

Subreddit confusion statistics indicate that certain communities are easier to identify than others. Both style and topic models do well in recognizing 
{\small\sf askscience:} classification accuracy for threads is as much as 97\%.
Communities that were most confusable are intuitively similar:
{\small\sf politics} and 
{\small\sf worldnews, askmen} and 
{\small\sf askwomen}.

\section{Community Feedback Correlation}
\label{sec:community-feedback}
\begin{figure}[t]
	\centering
	\includegraphics[width=0.4\textwidth]{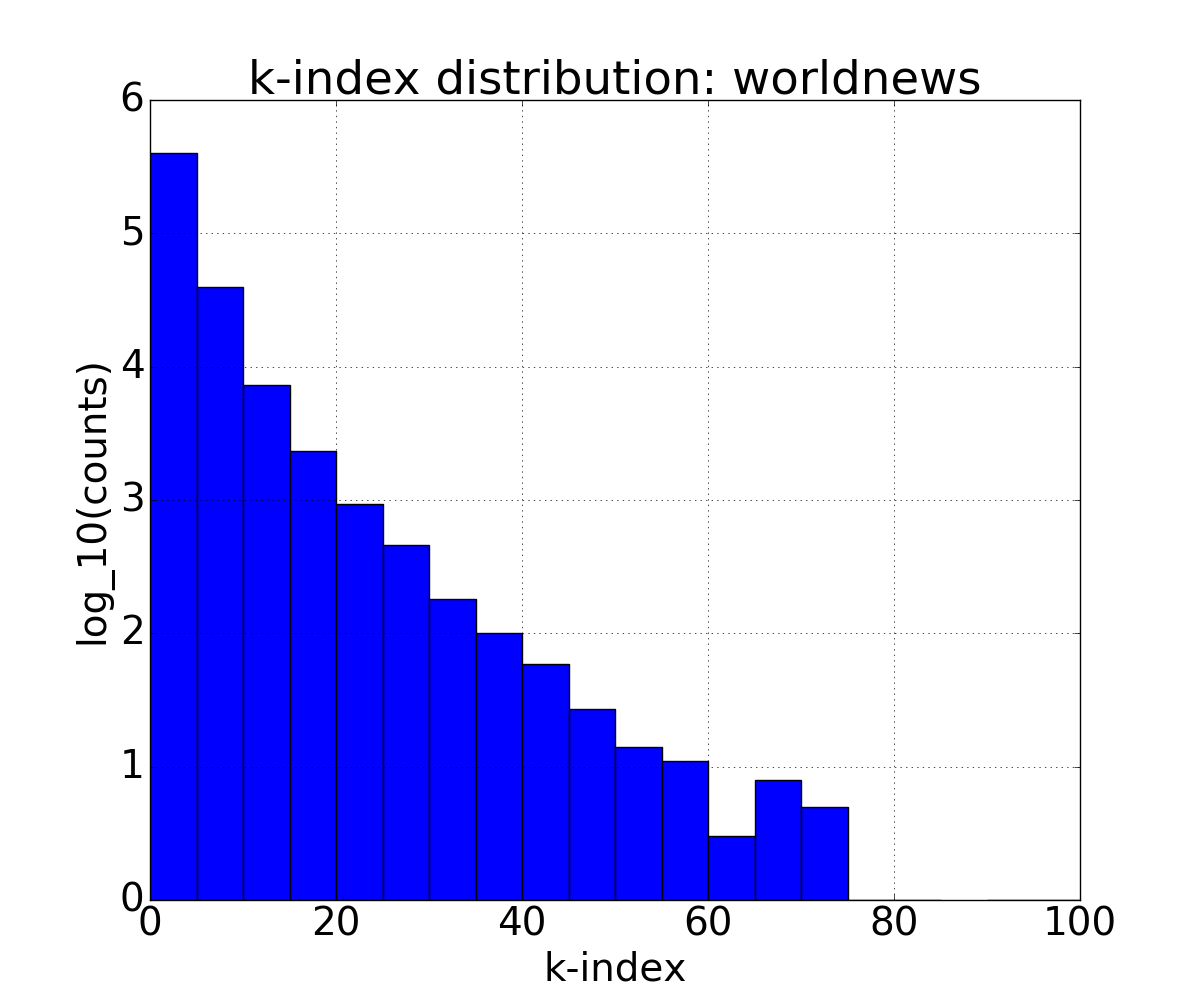}
	\caption{Distribution (log base 10 counts) of user k-index scores for the worldnews subreddit.}
	\label{fig:worldnews}
\end{figure}

In this section, we investigate whether the style and/or topic scores of a discussion or user are correlated with community response.
For thread-level feedback, we use karma points of the discussion thread itself; for the user-level feedback, we compute each user's subreddit-dependent k-index \cite{Jaech2015}, defined similarly to the well-known h-index \cite{Hirsch2005}. 
Specifically, a user's k-index $k_j$ in subreddit $j$ is the maximum integer $k$ such that the user has at least $k$ comments with karma greater than $k$ in that subreddit. User k-index scores have Zipfian distribution, as illustrated in Figure~\ref{fig:worldnews} for the {\small\sf worldnews} subreddit.

We compute a \emph{normalized} community similarity score $\tilde{s}_{i,j}=s_{i,j}-s_{i,m}$, where $s_{i,m}$ is the corresponding score from the subreddit 
{\small\sf merged\_others}. The correlation between $\tilde{s}_{i,j}$ and community feedback is  reported for three models in Table \ref{tab:correlation-post} for the thread level, and in Table \ref{tab:correlation-kindex} for the user level. 
On the thread level, the \verb+hyb-500.30+ style model consistently finds positive, statistically significant, correlation between the post's stylistic similarity score and its karma. This result suggests that language style adaptation does contribute to being well-received by the community. 
None of the other models explored in the previous section had this property, and for the topic models the correlation is mostly negative.
On the user level, \emph{all} correlations between a user's k-index and their style/topic match are statistically significant, though the \verb+hyb-500.30+ style model shows more positive correlation than other models. In both cases, the \verb+word_only+ model gives results between the style and topic models. The \verb+hyb-15k+ model has results that are similar to the \verb+word_only+ model, and the \verb+tag_only+ model has mostly negative correlation.

\begin{table}[]
\begin{tabular}{l|c|c|c} 
subreddit & hyb-500.30 & word\_only & topic-100  \\ \hline \hline
askmen & 0.392* & 0.222* & 0.055 \\
askscience & 0.321* & -0.110 & -0.166* \\
askwomen & 0.501* & 0.388* & 0.005 \\
atheism & 0.137* & -0.229* & -0.251 \\
chgmyview & 0.167* & -0.121* & -0.306* \\
fitness & 0.130* & 0.017 & -0.313* \\
politics & 0.533* & 0.341* & 0.011 \\
worldnews & 0.374* & 0.148* & -0.277*
\end{tabular}
\centering
\caption{Spearman rank correlation of thread $\tilde{s}_{i,j}$ with karma scores. (*) indicates statistical significance  ($p<0.05$).}
\label{tab:correlation-post}
\end{table}


\begin{table}[]
\begin{tabular}{l|c|c|c}
subreddit & hyb-500.30 & word\_only & topic-100  \\ \hline \hline
askmen & 0.402 & 0.215 & 0.167 \\
askscience & 0.343 & 0.106 & 0.042 \\
askwomen & 0.451 & 0.260 & 0.165 \\
atheism & 0.296 &0.024 &  0.107 \\
chgmyview & 0.446 & 0.020 & 0.091 \\
fitness & 0.309 & 0.286 & 0.127 \\
politics & 0.453 & 0.317 & 0.177 \\
worldnews & 0.421 & 0.330 & 0.166
\end{tabular}
\centering
\caption{Spearman rank correlation of authors' $\tilde{s}_{i,j}$ with their k-indices. All values are statistically significant ($p<0.05$).}
\label{tab:correlation-kindex}
\end{table}


Examining users' multi-community involvement, we also find that users with high k-indices tend to participate in fewer subreddits. Among relatively active users (having at least 100 comments), those with a max k-index of at least 100 participated in a median of 3 communities, while those with a max k-index of at most 5 participated in a median of 6 subreddits. Of the 42 users with max k-index of at least 100, only 4 achieve a k-index of at least 50 in one other community, and only 6 achieve a k-index of at least 20 in one other community. 
\section{Conclusion}
\label{sec:conclusion}
In this work, we use hybrid n-grams and topic models to characterize style and topic of language in online communities. Since communities center on a common theme, topic characteristics are reflected in language style, but we find that the best model for determining community identity uses very few words and mostly relies on POS patterns. Using Reddit's community response system (karma), we also show that discussions and users with higher community endorsement are more likely to match the language style of the community, where the language model that best classifies the community is also most correlated with community response. In addition, online users tend to have more positive community response when they specialize in fewer subreddits. These results have implications for detecting newcomers in a community and the popularity of posts, as well as for language generation.




\subsection*{Acknowledgments}
This paper is based on work supported by the DARPA DEFT Program. Views expressed are those of the authors and do not reflect the official policy or position of the Department of Defense or the U.S.\ Government. We thank the reviewers for their helpful feedback.

\bibliography{emnlp2016}
\bibliographystyle{emnlp2016}

\end{document}